\documentclass[10pt,twocolumn,letterpaper]{article}

\usepackage{cvpr}
\usepackage{times}
\usepackage{epsfig}
\usepackage{graphicx}
\usepackage{amsmath}
\usepackage{amssymb}
\usepackage{multirow}
\usepackage{makecell}
\usepackage{array}
\usepackage{tabularx}
\usepackage{authblk}
\usepackage{fancyhdr}

\usepackage[pagebackref=true,breaklinks=true,letterpaper=true,colorlinks,bookmarks=false]{hyperref}

\cvprfinalcopy 

\def\cvprPaperID{6645} 

\ifcvprfinal\fi
\begin{document}

\title{Exploring Spatial-Temporal Multi-Frequency Analysis for High-Fidelity and Temporal-Consistency Video Prediction}

\author[1,2]{Beibei Jin}
\author[1,2]{Yu Hu \thanks{Corresponding author: Yu Hu, huyu@ict.ac.cn. This work is supported in part by the National Key R\&D Program of China under grant No. 2018AAA0102701, in part by the Science and Technology on Space Intelligent Control Laboratory under grant No. HTKJ2019KL502003, and in part by the Innovation Project of Institute of Computing Technology, Chinese Academy of Sciences under grant No. 20186090.}}
\author[1,2]{Qiankun Tang}
\author[1,2]{Jingyu Niu}
\author[3]{Zhiping Shi}
\author[1,2]{Yinhe Han}
\author[1,2]{Xiaowei Li}
\affil[1]{Research Center for Intelligent Computing Systems, State Key Laboratory of Computer Architecture\\
Institute of Computing Technology, Chinese Academy of Sciences}
\affil[2]{University of Chinese Academy of Sciences \authorcr{\{\tt\small jinbeibei, huyu, tangqiankun, niujingyu17b, yinhes, lxw\}@ict.ac.cn}}
\affil[3]{Capital Normal University \authorcr{\tt\small shizp@cnu.edu.cn}}

\maketitle
\thispagestyle{empty}

\begin{abstract}
   Video prediction is a pixel-wise dense prediction task to infer future frames based on past frames. Missing appearance details and motion blur are still two major problems for current models, leading to image distortion and temporal inconsistency. We point out the necessity of exploring multi-frequency analysis to deal with the two problems. Inspired by the frequency band decomposition characteristic of Human Vision System (HVS), we propose a video prediction network based on multi-level wavelet analysis to uniformly deal with spatial and temporal information. Specifically, multi-level spatial discrete wavelet transform decomposes each video frame into anisotropic sub-bands with multiple frequencies, helping to enrich structural information and reserve fine details. On the other hand, multi-level temporal discrete wavelet transform which operates on time axis decomposes the frame sequence into sub-band groups of different frequencies to accurately capture multi-frequency motions under a fixed frame rate. Extensive experiments on diverse datasets demonstrate that our model shows significant improvements on fidelity and temporal consistency over the state-of-the-art works. Source code and videos are available at \url{https://github.com/Bei-Jin/STMFANet}.
\end{abstract}

\section{Introduction}
\begin{figure}
    \centering
    \includegraphics[width=3.2in]{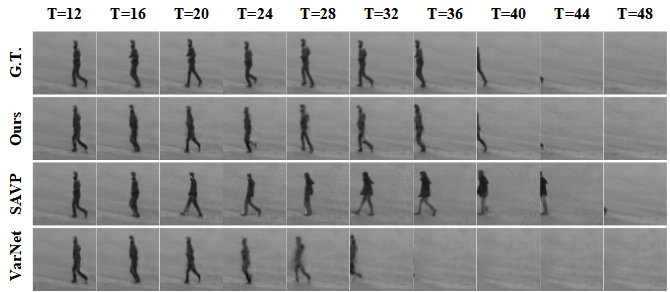}
   \caption{A comparison of long-term prediction on a KTH~\cite{Schuldt2004Recognizing} motion sequence. Our model generates predictions with higher fidelity and temporal consistency than the state-of-the-art methods, SAVP~\cite{lee2018stochastic} and VarNet~\cite{Jin2018Var}. In the other two methods' predictions, the person gradually blurs to distortion and runs out of the image too fast or too slowly, which is inconsistent to the ground truth.}
    \label{fig:gap}
\end{figure}
Unsupervised video prediction has attracted more and more attention in the research community and AI companies. It aims at predicting upcoming future frames based on the observation of previous frames. This looking-ahead ability has a broad application prospect on video surveillance~\cite{Elafi2016Unsupervised}, robotic systems~\cite{finn2017deep} and autonomous vehicles~\cite{Wei2010A}. However, building an accurate predictive model still remains challenging because it requires to master not only the visual abstraction model of different objects but also the evolution of various motions over time. Many recent deep learning methods~\cite{lee2018stochastic,Wei2018Novel,Tulyakov2017MoCoGAN,Byeon2017ContextVP,Villegas2017Learning,Villegas2017Decomposing,wang2019eidetic,kwon2019predicting} have brought about great development on the video prediction task. However, there still exists a clear gap between their predictions and the ground-truth (GT), as shown in Figure~\ref{fig:gap}. The predictions of the compared methods suffer from deficient retention of high-frequency details and insufficient use of motion information, resulting in distortion and temporal inconsistency:
\begin{figure}
    \centering
    \includegraphics[width=3.2in]{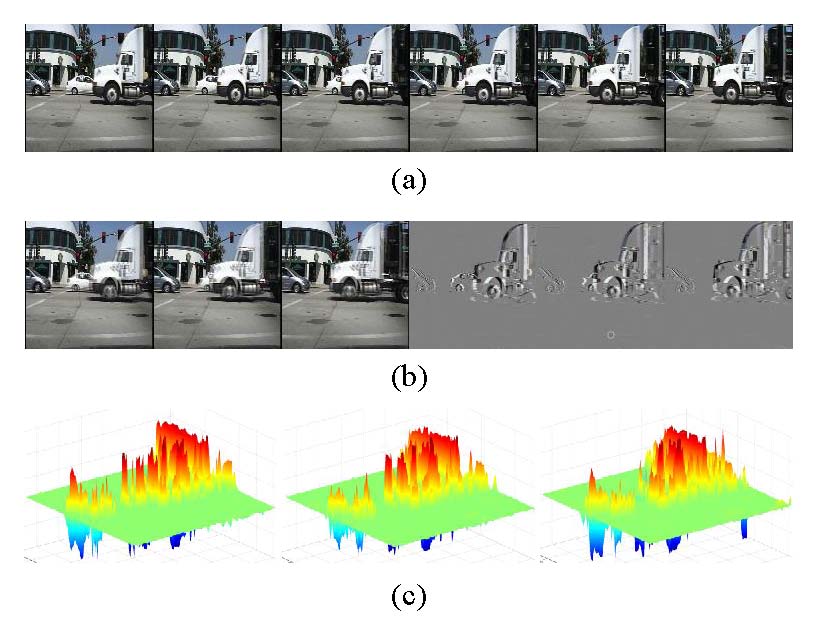}
   \caption{Discrete Wavelet Transform (DWT) on time axis can capture the different motion frequencies between the slower car and the faster truck. (a) is a video sequence with length six. DWT of (a) on time axis results in the sub-bands in (b). (c) is the heat maps of the right three sub-bands in (b), which can clearly show the difference between their movements.}
    \label{fig:T-DWT}
\end{figure}

\textbf{Loss of details.} Down-sampling is commonly adopted to enlarge the receptive field and extract global information, resulting in inevitable loss of high-frequency details. However, video prediction is a pixel-wise dense prediction problem. Sharp predictions would not be made without the assistance of fine details. Although dilated convolution can be employed to avoid using down-sampling, it has the problem of grid effect and is not friendly to small objects, which hinders the application to video prediction.

\textbf{Insufficient exploitation of temporal motions.} Dynamic scenes are composed of motions with more than one temporal frequency. In Figure~\ref{fig:T-DWT}, the lower temporal motion of the smaller car in the left and the faster temporal motion of the bigger truck in the right. They have different moving frequencies. However, previous methods usually process them one by one at a fixed frame rate. Although Recurrent Neural Networks (RNNs) are used to memorize dynamic dependencies, it has no ability to distinguish motions at different frequencies and cannot analyze time-frequency characteristics of temporal information.
\begin{figure}
    \centering
    \includegraphics[width=3.2in]{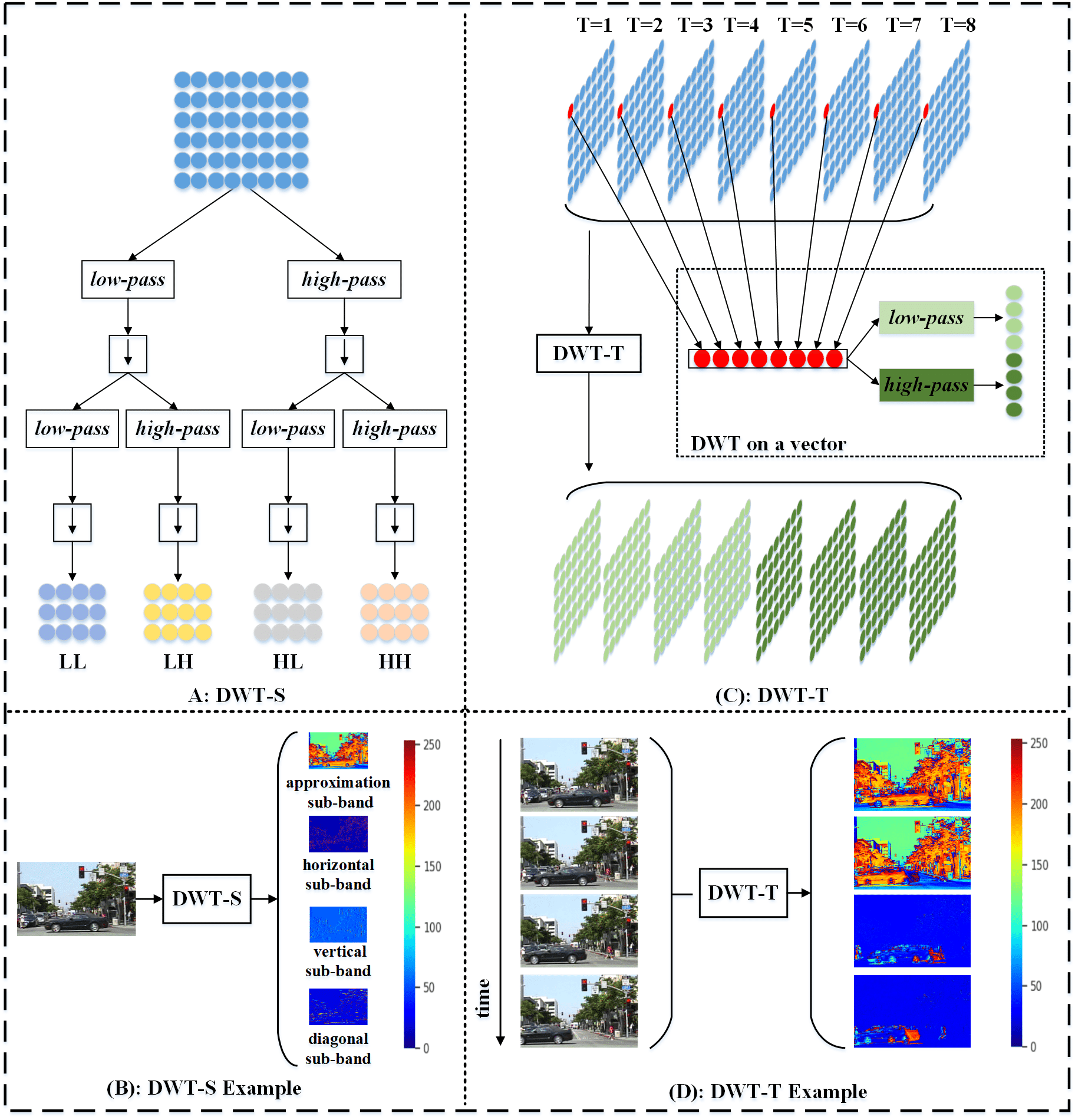}
    \caption{(A): Discrete Wavelet Transform in Spatial dimension (DWT-S) decomposes an image into one low frequency sub-band (LL) and three high frequency sub-bands of different directions (LH, HL, HH) which represent sub-bands of different directions (horizontal, vertical, diagonal). (B): An visualization example of (A). (C): Discrete Wavelet Transform in Temporal dimension (DWT-T) decomposes an image sequence into low frequency sub-bands and high frequency sub-bands on time axis. (D): An visualization example of (C). The sub-bands are visualized in heatmap style.}
    \label{fig:DWT-S-T}
\end{figure}

Therefore, it is necessary to introduce multi-frequency analysis into video prediction task. Biological studies~\cite{RTemporal, F1966Orientational} have shown that Human Visual System (HVS) exhibits multi-channel characteristics for spatial and temporal frequency information. The retinal images are decomposed into different frequency bands with approximately equal bandwidth on a logarithmic scale for processing~\cite{Mannos2003The}, which includes a low frequency band and multiple high frequency bands. Besides spatial dimension, there also is a similar frequency band decomposition in temporal dimension. These characteristics enable the Human Visual System (HVS) to process visual content with better discrimination of detailed information and motion information. Wavelet analysis~\cite{Chen2018DPW,Alemohammad2017High} is a spatial-scale (temporal-frequency) analysis method, which has the characteristic of multi-resolution (frequency) analysis and can well represent the local characteristics of spatial (temporal) frequency signal, which is very similar to HVS.

Discrete Wavelet Transform (DWT) is a common wavelet analysis method for image processing. As shown in Figure~\ref{fig:DWT-S-T}(B), the Discrete Wavelet Transform in Spatial dimension (DWT-S) ( Figure~\ref{fig:DWT-S-T}(A)) can decompose an image into one low frequency sub-band and three anisotropic high frequency sub-bands of different directions (horizontal, vertical, diagonal). Figure~\ref{fig:DWT-S-T}(D) shows the Discrete Wavelet Transform in Temporal dimension (DWT-T) (Figure~\ref{fig:DWT-S-T}(C)) decomposes a video sequence of length four into two high-frequency sub-bands and two low-frequency sub-bands on time axis. The frequency on time axis here can be viewed as how fast the pixels change with time, which is related to temporal motions. Inspired by the characteristics of HVS and wavelet transform, we propose to explore the multi-frequency analysis for high-fidelity and temporal-consistency video prediction. The main contributions are summarized as follows:
\begin{itemize}
    \item[1)] To the best of our knowledge, we are the first to propose a video prediction framework based on multi-frequency analysis that is trainable in an end-to-end manner.
    \item[2)] To strengthen the spatial details, we develop a multi-level Spatial Wavelet Analysis Module (S-WAM) to decompose each frame into one low-frequency approximation sub-band and three high-frequency anisotropic detail sub-bands. The high-frequency sub-bands represent the boundary details well and are in favor of sharpening the prediction details. Besides, multi-level decomposition forms a spatial frequency pyramid, helping to extract objects' features with multi scales.
    \item[3)] To fully exploit the multi-frequency temporal motions of objects in dynamic scenes, we employ a multi-level Temporal Wavelet Analysis Module (T-WAM) to decompose buffered video sequence into sub-bands with different time frequencies, promoting the description of multi-frequency motions and helping to comprehensively capture dynamic representations.
    \item[4)] Both quantitative and qualitative experiments on diverse datasets demonstrate a significant performance boost than the state-of-the-art. Ablation studies are made to show the generalization ability of our model and the evaluation of sub-modules.
\end{itemize}
\section{Related Work}
\label{Related Work}
\subsection{Video Generation and Video Prediction}
Video generation is to synthesize photo-realistic image sequences without the need to guarantee the fidelity of the results. It focuses on modeling the uncertainty of the dynamic development of video to produce results that may be inconsistent with the ground truth but reasonable. Differently, Video prediction is to perform deterministic image generation. It needs not only to focus on the per-frame visual quality, but also to master the internal temporal features to determine the most reliable development trend that is closest to the ground truth.

\textbf{Stochastic Video Generation.} Stochastic Video Generation models focus on handling the inherent uncertainty in predicting the future. They seek to generate multiple possible futures by incorporating stochastic models. Probabilistic latent variable models such as Variational Auto-Encoders (VAEs)~\cite{kingma2013auto,rezende2014stochastic} and Variational Recurrent Neural Networks (VRNNs)~\cite{chung2015recurrent} are the most commonly used structures. ~\cite{babaeizadeh2017stochastic} developed a stochastic variational video prediction (SV2P) method that predicted a different possible future for each sample of its latent variables, which was the first to provide effective stochastic multi-frame generation for real-world videos. SVG~\cite{denton2018stochastic} proposed a generation model that combined deterministic prediction of the next frame with stochastic latent variables, introducing a per-step latent variables model(SVG-FP) and a variant with a learned prior (SVG-LP). SAVP~\cite{lee2018stochastic} proposed a stochastic generation model combining VAEs and GANs.  ~\cite{castrejon2019improved} extended the VRNN formulation by proposing a hierarchical variant that used multiple levels of latents per timestep.

\textbf{High-fidelity Video Prediction.} High-fidelity Video Prediction models aim to produce naturalistic image sequences as close to the ground truth as possible. The main consideration is to minimize the reconstruction error between the true future frame and the generated future frame. Such models can be classified as direct prediction models~\cite{Srivastava2015Unsupervised,Wei2018Novel,wang2019eidetic,kwon2019predicting,Byeon2017ContextVP,Villegas2017Learning,Mathieu2015Deep,Villegas2017Decomposing,Jiang2017Super,liu2018dyan} and transformation-based prediction models~\cite{Zhou2016Learning,Vondrick2017Generating,Amersfoort2017Transformation,reda2018sdc}. Direct prediction models predict pixel values of future frames directly. They use a combination of forward neural network and recurrent neural network to encode spatial and temporal features, and then perform decoding to get the prediction with the corresponding decoding network. Generative adversarial networks (GANs) are often employed to make the predicted frames more realistic. Transformation-based prediction models aim at modeling the source of variability and operate in the space of transformations between frames. They focus on learning the transformation kernels between frames which are applied to the previous frames to synthesize the future frames indirectly.

Here, latent variables in stochastic video generation models is not considered in our model. Such models learn and sample from a space of possible futures to generate the subsequent frames. Although reasonable results can be generated by sampling different latent variables, there is no guarantee of consistency with the ground truth. Moreover, the quality of generation results vary from sample to sample, which is uncontrollable. This limits the application of such models in some practical tasks requiring a high degree of certainty, such as autonomous driving. We focus on high-fidelity video prediction, aiming to construct a prediction model to predict realistic future frame sequences as close to the ground truth as possible. To overcome the challenges of lack of details and motion blur, we propose to explore multi-frequency analysis based video prediction by incorporating wavelet transform with generative adversarial network.
\begin{figure*}
    \centering
    \includegraphics[width=6.9in]{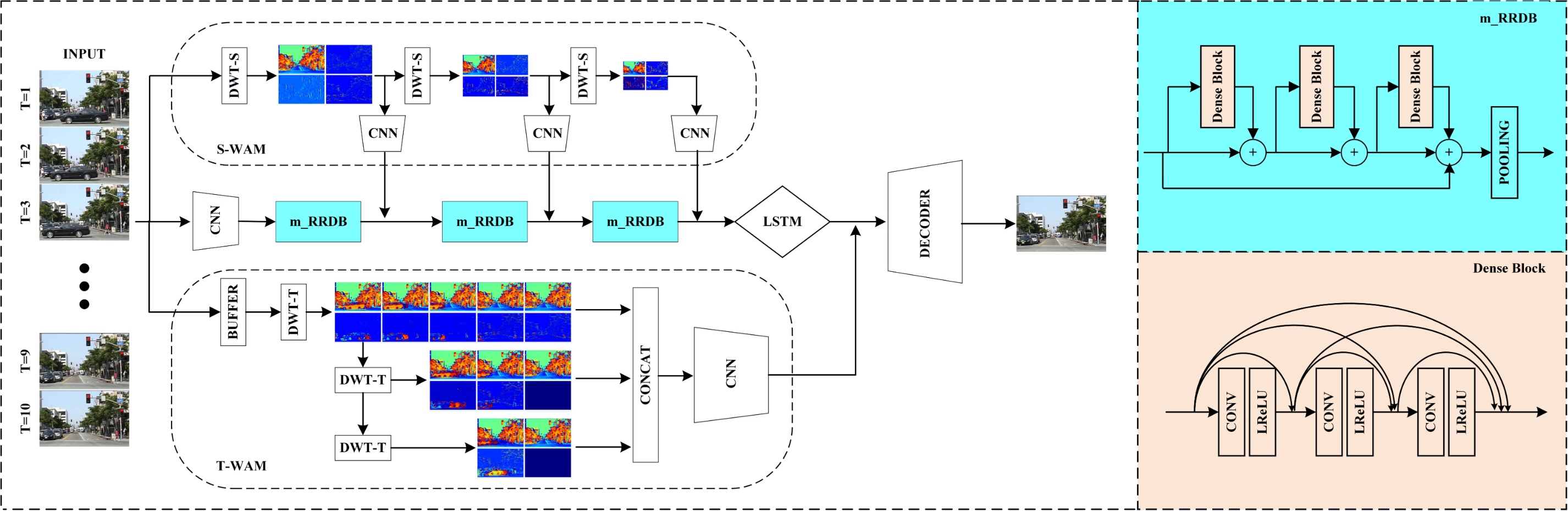}
   \caption{The pipeline architecture of our network. Note that the diagram takes the next frame prediction as an example. Multi-frame prediction can be done by feeding the predicted frame into the encoder network.}
    \label{fig:network}
\end{figure*}
\subsection{Wavelet Transform}
Wavelet Transform (WT) has been widely applied in image compression \cite{Chen2018DPW} and image reconstruction \cite{Huang2017Wavelet}. In image processing, Discrete Wavelet Transform (DWT) is often used. A fast implementation of it by using filter bank is proposed in~\cite{mallat1989theory}. The filter bank implementation of wavelets can be interpreted as computing the wavelet coefficients of a discrete set of child wavelets for a given mother wavelet. According to~\cite{mallat1989theory}, we illustrate the process of DWT on space axes of an image and DWT on time axis of a video sequence in Figure~\ref{fig:DWT-S-T}. Multi-level DWT can be done by repeating a similar process on a sub-band images. The multi-resolution (frequency) analysis of DWT is consistent with Human Visual System (HVS), which provides a biological basis for our approach. We recommend to refer to~\cite{mallat1989theory} to learn more about Discrete Wavelet Transform (DWT).
\section{Method}
\subsection{Problem Statement}
We aim to synthesize future frames of high fidelity and temporal consistency by observing several beginning frames. Let $X=\{x_i\}, (1\leq i \leq m)$ be the input of length $m$. $x_i\in\mathbb{R}^{H\times W\times C}$ represents the $i$th frame. H, W and C are the height, width and channel number. Let $Y=\{y_j\}, (1\leq j \leq n)$ represents the ground truth of future frame sequence of length $n$ and $\hat{Y}=\{\hat{y}_j\}, (1\leq j \leq n)$ represents the prediction of $Y$. The goal is to minimize the reconstruction error between $\hat{Y}$ and $Y$. We will take the next frame prediction as an example.
\subsection{Network Architecture}
We adopt generative adversarial network as the model structure. The Generator $G$ and discriminator $D$ are trained with competing goals: $G$ aims to predict frames that can fool $D$, while $D$ aims to distinguish whether the input samples are real (from the training dataset) or fake (from $G$).

Figure~\ref{fig:network} demonstrates the overall block diagram of the generator $G$ to predict frame $t+1$ at time step $t$. It follows an encoder-decoder architecture. The encoder aims to transform the input sequence into a hidden feature tensor, while the decoder is in charge of decoding the feature tensor to generate the prediction of the next frame. The encoder consists of three part: stem CNN-LSTM, cascaded Spatial Wavelet Analysis Modules (S-WAMs) and Temporal Analysis Module (T-WAM). The decoder is composed of deconvolution and up-sampling layers.

The stem encoder is a \emph{'CNN-LSTM'} structure. At each time step $t$ $(t\geq 1)$, the frame $x_t$ is passed through the stem network to extract multi-scale spatial information under different receptive fields. To pursue a better expression of appearance features, we refer to the Residual-in-Residual Dense Block (RRDB) proposed by~\cite{wang2018esrgan} in the design of our stem structure. It is a combination of multi-level residual network and dense connections. We make a modification: adding a down-sampling layer in each RRDB unit to reduce the size of feature maps.

To reserve more high-frequency spatial details, considering multi-resolution analysis of wavelet transform, we propose a Spatial Wavelet Analysis Module (S-WAM) to enhance the representation of high-frequency information. As illustrated in Figure \ref{fig:network}, S-WAM consists of two stages: Firstly, the input is decomposed into one low-frequency sub-band and three high-frequency detail sub-bands by DWT on Spatial dimension (DWT-S); Secondly, the sub-bands are fed into a shallow CNN to do further feature extraction and obtain consistent number of channels with the corresponding m\_RRDB unit. We cascade three S-WAMs to do multi-level wavelet analysis. The output of each level of S-WAM is added with the corresponding feature tensors of the m\_RRDB unit. The cascaded S-WAMs provide the compensation of details to the stem network under multiple frequencies, which promotes the prediction of fine details.

On the other side, to model the temporal multi-frequency motions in video sequences, we design a multi-level Temporal Wavelet Analysis Module (T-WAM) decomposing the sequence into sub-bands under different frequencies on time axis. In our experiments, we conduct multi-level DWT on temporal dimension (DWT-T) on the input sequence until the number of low-frequency sub-bands or high-frequency sub-bands equals two. We take three DWT-T as an example in Figure~\ref{fig:network}. Then we concatenate those sub-bands as the input of a CNN to extract features and adjust the size of feature maps. The output is fused with the historical information from LSTM cell to strengthen the ability to distinguish multi-frequency motions for the model. The fused feature tensors from the encoder network are fed to the decoder network to generate the prediction of the next frame. We conduct a discriminator network as~\cite{Mathieu2015Deep} and train the discriminator to classify the input $[X,\hat{Y}]$ into class $0$ and the input $[X,Y]$ into class $1$.
\subsection{Loss Function}
We adopt multi-module losses which consists of the image domain loss and the adversarial loss.

\textbf{Image Domain Loss.} We combine $\mathcal{L}_2$ loss with the Gradient Difference Loss (GDL)~\cite{Mathieu2015Deep} as the image domain loss:
\begin{equation}
    \mathcal{L}_{img}(Y,\hat{Y})=\mathcal{L}_2(Y,\hat{Y})+\mathcal{L}_{gdl}(Y,\hat{Y}).
\end{equation}
\begin{equation}
    \mathcal{L}_{2}(Y,\hat{Y})=\vert\vert(Y-\hat{Y})\vert\vert_2^2=\sum_{i=1}^{n}{\left \|{(y_i-\hat{y}_i)}\right \|}_2^2.
\end{equation}
\begin{eqnarray}
\begin{aligned}
    \mathcal{L}_{gdl}(Y,\hat{Y})=&\sum_{i=1}^n\sum_{i,j}{\big|{{\vert{y_{i,j}-y_{i-1,j}}\vert}-{\vert{\hat{y}_{i,j}-\hat{y}_{i-1,j}}\vert}}\big|}^\alpha\\&+{\big|{{\vert{y_{i,j-1}-y_{i,j}}\vert}-{\vert{\hat{y}_{i,j-1}-\hat{y}_{i,j}}\vert}}\big|}^\alpha,
\label{eq:GDLloss}
\end{aligned}
\end{eqnarray}
where $\alpha$ is an integer greater or equal to $1$, and $\vert . \vert$ is the operation of absolute value function.

\textbf{Adversarial Loss.} Adversarial training involves a generator G and a discriminator D, where D learns to distinguish whether the frame sequence is from the real dataset or produced by G. The two networks are trained alternately, thus improving until D can no longer discriminate the frame sequence generated by G. In our model, the prediction model is regarded as a generator. We formulate the adversarial loss on the discriminator D as:
\begin{equation}
    \mathcal{L}_{D}^A=-log D([X,Y])-log(1-D(X,\hat{Y})),
\end{equation}
and the adversarial loss for the generator G as:
\begin{equation}
   \mathcal{L}_{G}^A=-log D([X,\hat{Y}]).
\end{equation}

Hence, we combine the losses previously defined for our generator model with different weights:
\begin{eqnarray}
\begin{aligned}
    \mathcal{L}_{G}=\lambda_1\mathcal{L}_{img}+\lambda_2\mathcal{L}_{G}^A,
\label{eq:L_all}
\end{aligned}
\end{eqnarray}
where $\lambda_1$ and $\lambda_2$ are hyper-parameters to trade off between these distinct losses.
\begin{figure*}[]
    \centering
    \includegraphics[width=6.9in]{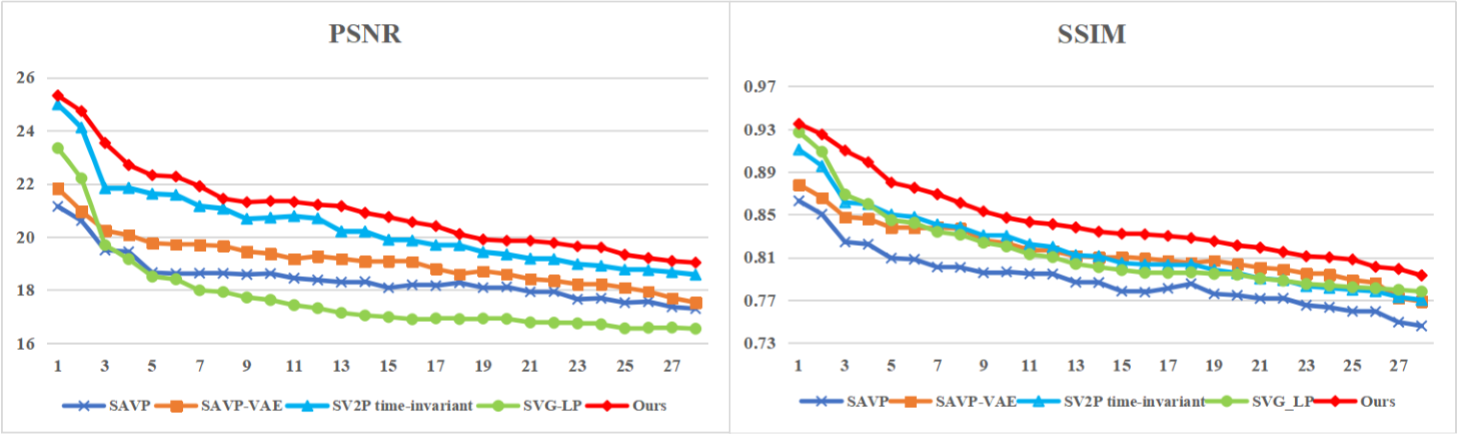}
   \caption{Quantitative comparison of different prediction models on BAIR datasets. Higher values for both PSNR and SSIM indicate better performance.}
    \label{fig:BAIR-PSNR-SSIM}
\end{figure*}
\section{Experiments}
\subsection{Experiment Setup}
\textbf{Datasets.} We perform experiments on diverse datasets widely used to evaluate video prediction models. The KTH dataset~\cite{Schuldt2004Recognizing} contains 6 types of actions from 25 persons. We use person 1-16 for training and 17-25 for testing. Models are trained to predict next 10 frames based on the observation of previous 10 frames. The prediction range of testing is extended to 20 or 40 frames. The hyper parameters in the loss function on KTH dataset are: $\lambda_1=1$ and $\lambda_2=0.01$. The BAIR dataset~\cite{ebert2017self} consists of a random moving robotic arm that pushes objects on a table. This dataset is particularly challenging due to the high stochasticity of the arm movements and the diversity of the background. We follow the setup in ~\cite{lee2018stochastic} and the hyper parameters in the loss function on the BAIR dataset are: $\lambda_1=1$ and $\lambda_2=0.001$. In addition, following the experiments settings in~\cite{liang2017dual}, we validate the generalization ability of our models on the car-mounted camera datasets (train: KITTI dataset~\cite{geiger2013vision}, test:Caltech Pedestrian dataset~\cite{Piotr2012Pedestrian}). The hyper parameters are: $\lambda_1=1$ and $\lambda_2=0.001$.

\textbf{Metrics.} Quantitative evaluation of the the accuracy is performed based on Peak Signal-to-Noise Ratio (PSNR) and Structural Similarity Index Measure (SSIM) metrics~\cite{Zhou2004Image}. Higher values indicate better results. To measure the realism of predicted results, we employ the metric of Learned Perceptual Image Patch Similarity (LPIPS)~\cite{zhang2018unreasonable}. Fr¨¦chet Video Distance (FVD)~\cite{unterthiner2018towards} is also adopted to evaluate the distribution over entire videos.

\begin{table}[h]
\caption{The average comparison results over predicted $20$ time steps $(10 \rightarrow 20)$ and $40$ time steps $(10 \rightarrow 40)$ based on $10$ time steps on the KTH dataset. The best results under each metric are marked in bold.}
\footnotesize
\begin{tabularx}{3.4in}{X|XXXXXXX}
\hline
\multicolumn{1}{l}{\multirow{3}{*}{Method}} & \multicolumn{6}{c}{KTH} \\
\multicolumn{1}{l}{} & \multicolumn{3}{c}{$10 \rightarrow 20$}  & \multicolumn{3}{c}{$10 \rightarrow 40$} \\
\multicolumn{1}{l}{} & PSNR & SSIM  & LPIPS & PSNR & SSIM & LPIPS \\
\hline
\multicolumn{1}{l}{MCNET~\cite{Villegas2017Decomposing}} & \multicolumn{1}{l}{25.95} & 0.804 & \multicolumn{1}{c}{-} & 23.89 & 0.73 & \multicolumn{1}{c}{-} \\
\multicolumn{1}{l}{fRNN~\cite{oliu2018folded}} & 26.12  & 0.771 & \multicolumn{1}{c}{-} & 23.77 & 0.678 & \multicolumn{1}{c}{-}\\
\multicolumn{1}{l}{PredRNN~\cite{wang2017predrnn}}  & 27.55 & 0.839  & \multicolumn{1}{c}{-} & 24.16 & 0.703 & \multicolumn{1}{c}{-}\\
\multicolumn{1}{l}{PredRNN++~\cite{Wang2018PredRNN++}}  & 28.47  & 0.865  & \multicolumn{1}{c}{-} & 25.21 & 0.741 & \multicolumn{1}{c}{-} \\
\multicolumn{1}{l}{VarNet~\cite{Jin2018Var}} & 28.48 & 0.843 & \multicolumn{1}{c}{-} & 25.37 & 0.739 & \multicolumn{1}{c}{-} \\
\multicolumn{1}{l}{E3D-LSTM~\cite{wang2019eidetic}} & 29.31 & 0.879  & \multicolumn{1}{c}{-} & 27.24 & 0.810 & \multicolumn{1}{c}{-}\\
\multicolumn{1}{l}{MSNET~\cite{LeeMutual}} & 27.08  & 0.876 & \multicolumn{1}{c}{-} & \multicolumn{1}{c}{-} & \multicolumn{1}{c}{-} & \multicolumn{1}{c}{-} \\
\hline
\multicolumn{1}{l}{SAVP~\cite{lee2018stochastic}} & 25.38 & 0.746 & \multicolumn{1}{l}{9.37} & 23.97 & 0.701 & 13.26 \\
\multicolumn{1}{l}{SAVP-VAE~\cite{lee2018stochastic}} & 27.77 & 0.852 &  \textbf{8.36} & 26.18 & 0.811 & \textbf{11.33} \\
\multicolumn{1}{l}{SV2P time-invariant~\cite{babaeizadeh2017stochastic}} & 27.56 & 0.826  & 17.92 & 25.92 & 0.778 & 25.21 \\
\multicolumn{1}{l}{SV2P time-variant~\cite{babaeizadeh2017stochastic}} & 27.79 & 0.838 & 15.04 & 26.12 & 0.789 & 22.48 \\
\hline
\multicolumn{1}{l}{Ours} & \textbf{29.85} & \textbf{0.893} & 11.81  & \textbf{27.56} & \textbf{0.851}  & 14.13  \\
\multicolumn{1}{l}{Ours (w/o S-WAM)} & 29.13 & 0.872 & 12.33  & 26.42 & 0.805  & 16.06  \\
\multicolumn{1}{l}{Ours (w/o T-WAM)} & 28.57 & 0.839 & 15.16 & 26.08 & 0.782  & 17.45  \\
\multicolumn{1}{l}{Ours (w/o WAM)} & 27.37 & 0.821 & 18.31 & 24.03 & 0.721  & 20.07  \\
\hline
\end{tabularx}
\label{Table:KTH results}
\end{table}
\begin{table}[h]
\centering
\caption{Quantitative evaluation of different methods on the BAIR dataset. The metrics are averaged over the predicted frames. The
best results under each metric are marked in bold.}
\begin{tabular}{l|ccc}
\hline
\multirow{2}{*}{Method} & \multicolumn{3}{c}{BAIR} \\
& PSNR & SSIM  & LPIPS \\
\hline
SAVP~\cite{lee2018stochastic} & 18.42 & 0.789 & 6.34 \\

SAVP-VAE~\cite{lee2018stochastic} & 19.09 & 0.815 & 6.22 \\

SV2P time-invariant~\cite{babaeizadeh2017stochastic} & 20.36 & 0.817 & 9.14 \\

SVG-LP~\cite{denton2018stochastic} & 17.72 & 0.815 & 6.03 \\

Improved VRNN~\cite{castrejon2019improved}  & -  & 0.822 & \textbf{5.50} \\
\hline
Ours & \textbf{21.02} & \textbf{0.844} & 9.36 \\
Ours (w/o S-WAM) & 20.22 & 0.825 & 11.23 \\
Ours (w/o T-WAM) & 19.87 & 0.819 & 11.72 \\
Ours (w/o WAM) & 18.15 & 0.784 & 13.13 \\
\hline
\end{tabular}
\label{Table:BAIR average results}
\end{table}
\subsection{Quantitative Evaluation}
The results of methods~\cite{Villegas2017Decomposing, oliu2018folded, wang2017predrnn, Wang2018PredRNN++, Jin2018Var, wang2019eidetic, LeeMutual, castrejon2019improved} are reported in the reference papers~\cite{wang2019eidetic, Jin2018Var, LeeMutual, castrejon2019improved}. For the models~\cite{lee2018stochastic, babaeizadeh2017stochastic, denton2018stochastic}, we generate the results by running the pre-trained models the authors reported online. Table~\ref{Table:KTH results} reports quantitative comparison on the KTH dataset. We can see that our model achieves the best result on PSNR and SSIM in terms of prediction for both future 20 frames and 40 frames, which indicates that our results are more consistent with the ground truth. However, on LPIPS, SAVP and its variants SAVP-VAE perform better than us. We analyze that the introduction of latent variables in the stochastic generation methods focuses more on the visual quality of the generated results and less on the consistency with ground truth. Nevertheless, our model focuses more on fidelity and temporal consistency with the original sequences, which is in line with our original intention.

Figure~\ref{fig:BAIR-PSNR-SSIM} illustrates the per-frame quantitative comparison on the BAIR dataset. We also calculate the average results in Table~\ref{Table:BAIR average results}. In consistent with the result on KTH dataset, we obtain the best PSNR and SSIM among the reported methods. While the Improved VRNN~\cite{castrejon2019improved} achieves the highest on LPIPS. Because of the high stochasticity of the BAIR dataset, it is challenging to maintain fidelity and temporal consistency while making good visual effects. Besides frame-wise comparison, we adopt FVD (Fr¨¦chet video Distance)~\cite{unterthiner2018towards} to evaluate the distribution over entire sequences. As shown in Table~\ref{table:FVD}, our FVD results are competitive to other methods on both datasets, which shows the consistency of the distribution of the predicted sequences.
\begin{table}
\centering
\caption{FVD (the smaller the better) evaluation on KTH and BAIR dataset. Baselines did not evaluate on KITTI and CalTech Pedestrian.}
\label{table:FVD}
\resizebox{7cm}{0.6cm}{
\begin{tabular}{ccccc}
\hline
Dataset & SVG-FP & SV2P & SAVP & Ours \\ \hline
KTH & 208.4~\cite{unterthiner2018towards} & 136.8~\cite{unterthiner2018towards} & 78.0~\cite{unterthiner2018towards} & \bfseries{72.3}\\
BAIR & 315.5~\cite{unterthiner2018towards} & 262.5~\cite{unterthiner2018towards} & \bfseries{116.4}~\cite{unterthiner2018towards} & 159.6\\ \hline
\end{tabular}
}
\end{table}

\begin{figure*}
    \centering
    \includegraphics[width=6.8in]{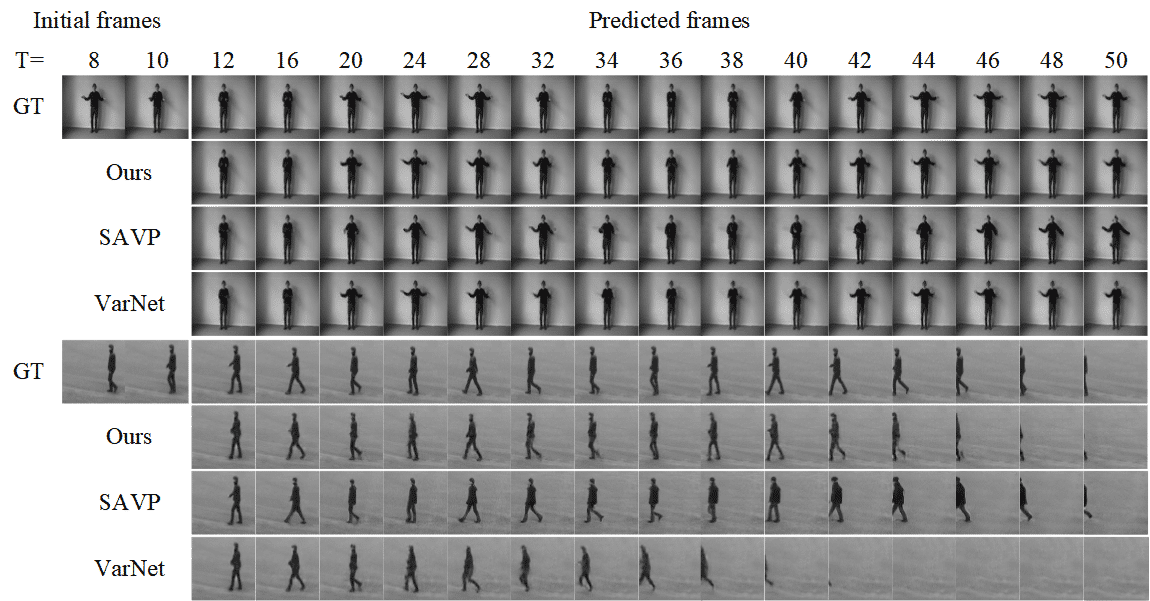}
   \caption{The prediction visualization of future 40 time steps based on the 10 frames on the KTH dataset.}
    \label{fig:KTH visual}
\end{figure*}
\begin{figure*}[]
    \centering
    \includegraphics[width=6.8in]{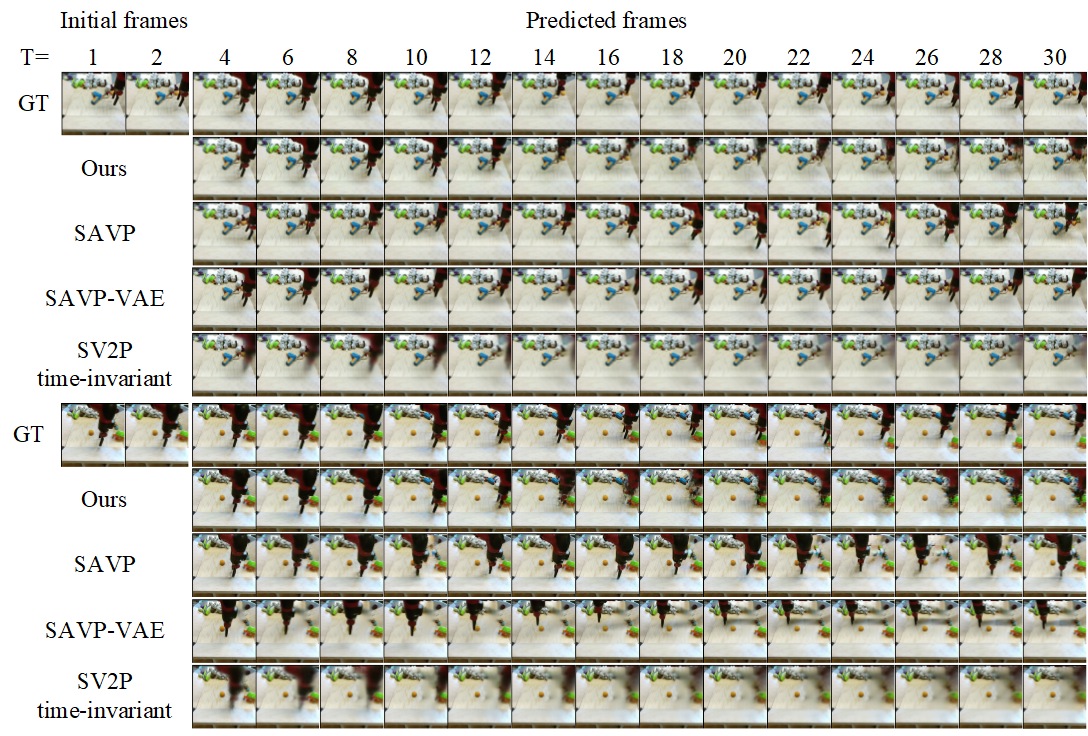}
   \caption{The prediction visualization comparison on the BAIR action free dataset. Our model predicts more consistent results to the ground truth.}
    \label{fig:BAIR visual}
\end{figure*}
\begin{figure*}
    \centering
    \includegraphics[width=6.6in]{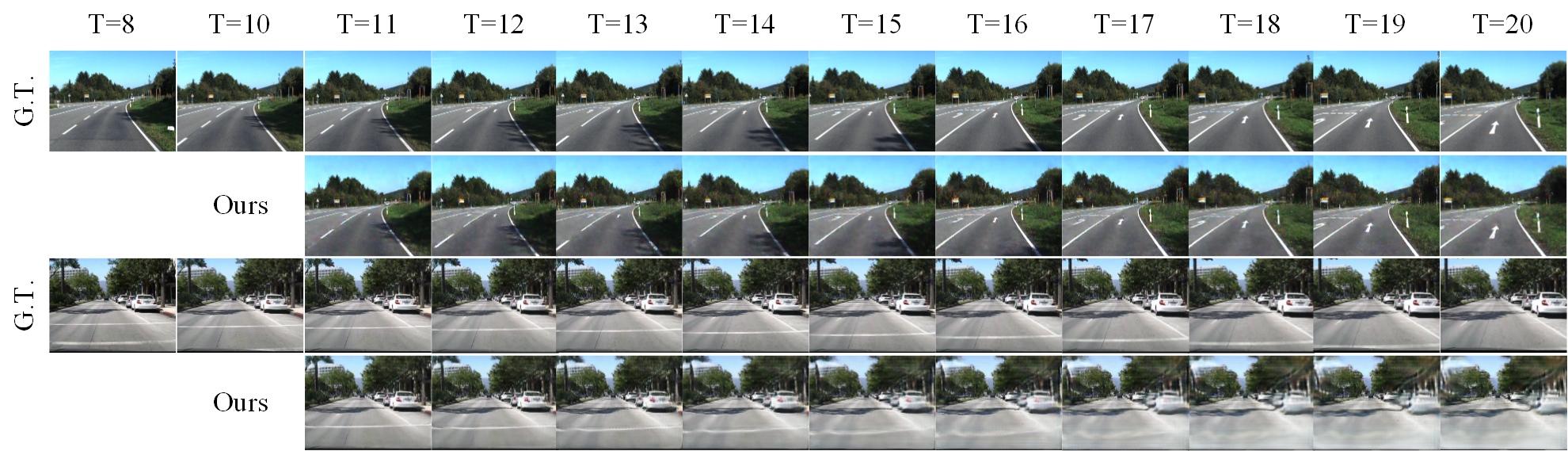}
   \caption{Visualization examples on KITTI dataset (the first group) and CalTech Pedestrian dataset (the second group).}
    \label{fig:generalization test}
\end{figure*}
\subsection{Qualitative Evaluation}
We report visualization examples on KTH dataset and BAIR datasets in Figure~\ref{fig:KTH visual} and ~\ref{fig:BAIR visual}. The first row is the ground truth, where the initial frames represent the input frames. Our model makes more accurate predictions while maintaining more details of the arms in the handclapping example in first group of Figure~\ref{fig:KTH visual}. Meanwhile, we predict a walking sequence that is more consistent with the ground truth in the second group of Figure~\ref{fig:KTH visual}, while for other methods, the person in the image walks out of the scene too quickly (VarNet) or two slowly (SAVP and SV2P time-invariant).
For the predictions on BAIR dataset, we are also the most consistent. Though the stochastic generation methods seem to generate more clear results, they are very different from the moving trajectories of the real sequence. This again confirms our belief that introducing more stochasticity in models will sacrifice fidelity. From the experiment results above, we can see that the multi-frequency analysis of discrete wavelet transform does help models to retain more detail information as well as temporal motion information.
\begin{table}[]
\setlength{\abovecaptionskip}{0pt}%
\setlength{\belowcaptionskip}{20pt}%
\small
\caption{Evaluation of Next frame prediction on the CalTech Pedestrian dataset after trained on the KITTI dataset. All models are trained by observing 10 frames.}
    \centering
    \begin{tabular}{l|ccccc}
    \hline
    Method & PSNR & SSIM  & LPIPS & \#param\\
    \hline
    PredNet~\cite{lotterdeep} & 27.6 & 0.905 & 7.47 & 6.9M \\
    ContextVP~\cite{Byeon2017ContextVP} & 28.7 & 0.921 & 6.03 & 8.6M\\
    DVF~\cite{LiuVideo} & 26.2 & 0.897 & 5.57 & 8.9M\\
    Dual Motion GAN~\cite{liang2017dual} & - & 0.899 & - & -\\
    CtrlGen~\cite{hao2018controllable} & 26.5 & 0.900 & 6.38 & - \\
    DPG~\cite{gao2019disentangling}  & 28.2 & 0.923 & \textbf{5.04} & - \\
    Cycle GAN~\cite{kwon2019predicting} & \textbf{29.2} & 0.830 & - & - \\
    \hline
    Ours & 29.1 & \textbf{0.927} & 5.89 & 7.6M \\
    Ours (w/o S-WAM) & 28.6 & 0.919 & 6.90 & 7.2M \\
    Ours (w/o T-WAM) & 28.1 & 0.903 & 7.56 & 7.3M \\
    Ours (w/o WAM) & 26.8 & 0.897 & 7.89 & 6.9M \\
    \hline
    \end{tabular}
    \label{Table:ablation study}
\end{table}
\subsection{Ablation Study}
\textbf{Evaluation of generalization ability.} Consistent with the previous works to evaluate the generalization ability, we test our model on the Caltech Pedestrian dataset after trained on KITTI dataset in Table~\ref{Table:ablation study}. We achieve the state-of-the-art performance. Figure~\ref{fig:generalization test} shows the visualization examples on KITTI dataset (the first group) and Caltech Pedestrian dataset (the second group). We can see that our model predicts clearly the evolution of driving lines and the cars. The results remain consistent with the ground truth, which verifies the good generalization ability of the model. Besides, we report the number of model's parameters in Table~\ref{Table:ablation study}. Compared to ContextVP~\cite{Byeon2017ContextVP} and DVF~\cite{LiuVideo}, our model achieves better results with fewer parameters.

\begin{figure}[t]
\begin{center}
\setlength{\abovecaptionskip}{-0.2cm}
\setlength{\belowcaptionskip}{-0.2cm}
   \includegraphics[width=3.3in]{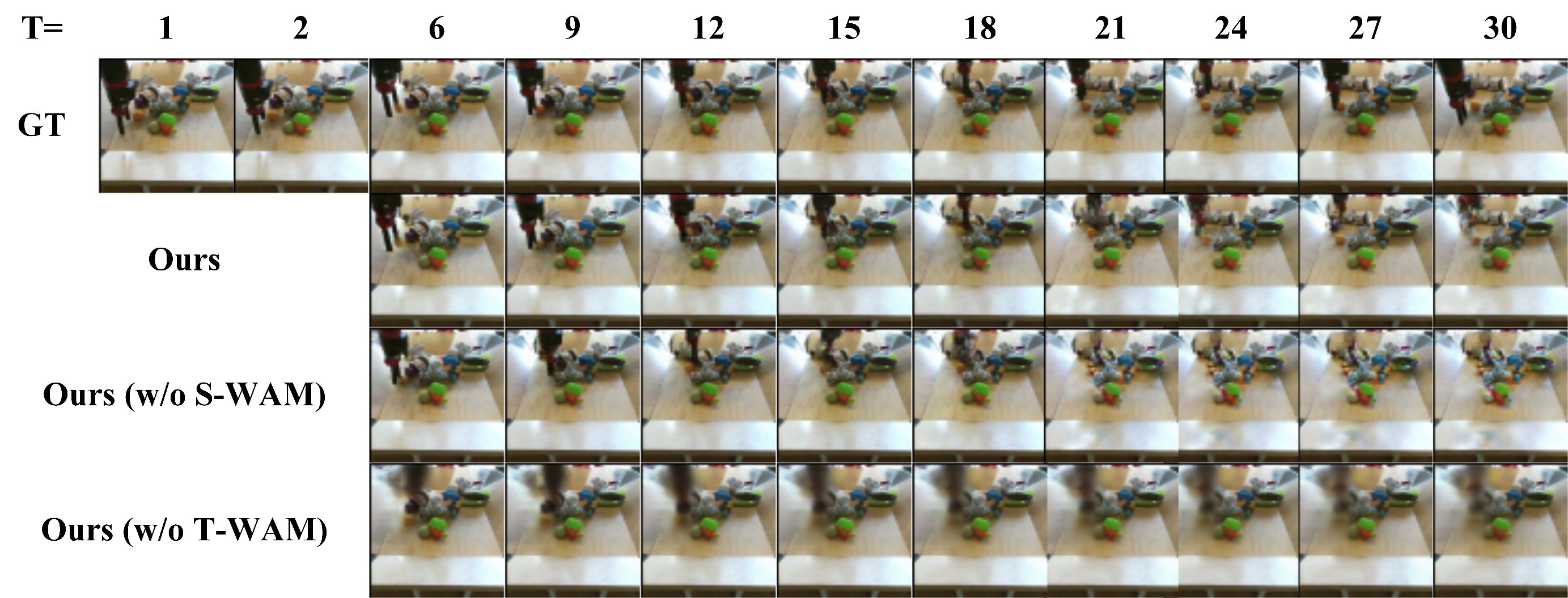}
\end{center}
   \caption{A BAIR Failure case. Best viewed by zooming.}
\label{fig:failure cases}
\end{figure}

\textbf{Evaluation of sub-modules.} To assess the impact of each sub-module, we do ablation studies in the absence of S-WAM and/or T-WAM. Results suggest that sub-modules, S-WAM and T-WAM, have both contributed to improving the prediction effect. Specifically the model without S-WAM gains more than the model without T-WAM. The visualization in Figure~\ref{fig:failure cases} is consistent. We analyze that the temporal motion information is of vital importance in the long-term prediction, especially for long-term prediction. Improving the expression of multi-frequency motion information in the model is the basis for making predictions with high-fidelity and temporal-consistency.

\textbf{Analysis of failure cases.} As shown in Figure~\ref{fig:failure cases}, for beginning motion under certain historical dependence, {\bfseries{Ours}} model predicts accurately. Since an abrupt movement occurs (18th - 21th frame), predictions of robotic manipulator become incorrect. BAIR is indeed of high stochasticity due to the action variability. Our T-WAM module extracts the transient features of the sequence, in addition to decomposing the input into sub-band groups of different frequencies to accurately capture multi-frequency motions. However, maintaining high fidelity to accommodate abrupt motions is challenging, even for stochastic models, unless the corresponding action priors are added.
\section{Conclusion}
We discuss the issues of missing details and ignoring temporal multi-scale motions in current prediction models, which always lead to blurry results. Inspired by the mechanism in Human Visual System (HVS), we explore a video prediction network based on multi-frequency analysis, integrating spatial-temporal wavelet transform and generative adversarial network. The Spatial Wavelet Analysis Module (S-WAM) is proposed to reserve more details through multi-level decomposition of each frame. The Temporal Wavelet Analysis Module (T-WAM) is proposed to exploit the temporal motions through multi-level decomposition of video sequences on time axis. Extensive experiments demonstrate the superiority of our method over the latest methods.
{\small
\bibliographystyle{ieee_fullname}
\bibliography{egbib}
}

\end{document}